# AIQ: Measuring Intelligence of Business AI Software

Moshe BenBassat* (moshe.benbassat@plataine.com)
Arison School of Business, Interdisciplinary Center (IDC), Herzliya, Israel

If you wish to read more about my AI work and earlier publications, please visit www.moshebenbassat.com, contact me at: moshe.benbassat@plataine.com and follow me on LinkedIn: www.linkedin.com/in/moshe-benbassat-AI

## Abstract
Focusing on Business AI, this article introduces the **AIQ quadrant** that enables us to measure AI for business applications in a relative comparative manner, i.e. to judge that software A has more or less intelligence than software B. Recognizing that the goal of Business software is to maximize value in terms of business results, the dimensions of the quadrant are the key factors that determine the business value of AI software: **Level of Output Quality ("Smartness") and Level of Automation.** The use of the quadrant is illustrated by several software solutions to support the real-life business challenge of field service scheduling. The role of machine learning and conversational digital assistants in increasing the business value are also discussed and illustrated with a recent integration of existing intelligent digital assistants for factory floor decision-making with the new version of Google Glass. Such hands-free AI solutions elevate the AIQ level to its ultimate position.

**\*See bio at end of article**

## Introduction
As soon as Artificial Intelligence (AI) was contemplated, the need to define and measure it came up and is best represented by the well-known Turing test that offers a binary answer: YES-AI, NOT-AI. Considering that in today's world, AI existence is no longer questioned, at least in certain areas, it makes sense to look for more refined measures to assess how intelligent a given smart machine is. Focusing on Business AI, this article introduces the **AIQ quadrant that** enables us to measure AI for business applications in a relative comparative manner, i.e. to judge that software A has more or less intelligence than software B.

The goal of Business software is clear: maximize value in terms of business results; **revenue growth, cost reduction, market share and competitive differentiators;** driven also by the factor of **user acceptance**. (Higher user acceptance ensures that potential value is actually realized). The sophistication of the AI algorithms is only the enabler. What counts is the business value the software creates which is primarily derived from two sources: (1) **"Smartness"**- the quality of the decision the AI software generates, and (2) the level of **automation**. Accordingly, two main dimensions are proposed, which together, establish a **quadrant** for positioning and comparing AI software products:

1. **Level of Output Quality ("Smartness"):** Smart machines are expected to be smart. That is: automatically produce high quality solutions for problem solving and decision-making challenges, e.g. high accuracy prediction that a customer will churn, or equipment will fail, as well as higher schedule quality for manufacturing orders or field service calls.
2. **Level of Automation:** Is the software capable of automatically solving complex large-scale real-life business problems like building a full day service schedule in just one click? Or is it only capable of solving small segments of a large business problem guided by the user, e.g. user decides in which order to schedule the jobs, and the software schedules automatically one job at a time. Or is the software just a calculating machine fully guided step by step by the user, like we all do with straightforward Excel?

These concepts are illustrated below for the real-life business challenge of field service scheduling along with corresponding software products that exist in the market.

Finally, the emphasis so far in AI work has been on proving machine's ability to outperform human. With business results as the focus of Business AI, the question is not whether machine is smarter than human or the opposite, but rather how to create a man-machine team that performs better than any individual team member alone. The emphasis should be on identifying and allocating to each team member those tasks in which he has relative advantage over the other; meaning the focus should shift from what machines can do to what machines should do. This way they amplify each other's abilities leading to higher business value. These points, almost verbatim, first appeared in 1980 in an IEEE article that
described the MEDAS system we developed for Emergency and Critical Care and Space Medicine [1].

### Field Service Scheduling - AI Case Study
Sam is the scheduler of a field service operation. Every morning he is faced with hundreds/thousands of customer-orders and needs to find an optimal (or just 'good' quality) solution for the **W6** challenge of: **Who** (Technician) performs **What** job, for **Whom** (customer), **With** what (tools and spare parts), **When** (time slot) and **Where** [2]**.** Note that the sequence of jobs for each technician also determines the **route** he/she will travel. Schedule quality is measured by some mix of Key Performance Indicators (KPIs), including: percentage of compliance with SLAs (Service Level Agreements), total number of jobs per day (productivity), total travel miles (cost), total hours of overtime (cost), and more. The scheduler's job does not end with the morning schedule. Numerous events throughout the day disturb the schedule, e.g. new emergency jobs, jobs taking longer than planned, customers not at home, traffic delays, and these require on-going changes to the schedule **to accommodate the dynamics of the day** in a cost-effective manner.

The number of possible solutions for this scheduling problem grows exponentially as the volume of the customer-orders-input grows, and there is no algorithm that ensures converging in practical time to a mathematically optimal solution. This, however, does not mean we should do nothing. All it means is that we should compromise on the algorithm's goals and requirements and build around it **software products that produce sufficient business value that will make it worthwhile for people to pay for it and use it. (**The complexity of finding an optimal solution- even for moderate size service companies – is as high as playing chess or GO; actually higher).

The discussion below is based on decades of experience of developing AI solutions for a variety of **real-life complex** scheduling challenges including scheduling the Israeli air-force and ground forces, oil-refineries, and training facilities [2]. By the 1990's, our **W6 Service Scheduler** was already automatically producing schedules for large-scale field service operations with schedule quality far above that of any human being. It evolved into **ClickSchedule** from ClickSoftware which by now (2018) is approaching daily scheduling of roughly 1,000,000 Field Engineers (FE) for many of the world's largest service providers. Assuming that each field engineer delivers on average 3 to 3.5 jobs per day, and works roughly 210 field days per year, this means that, over a period of a year, ClickSchedule touches the life of about **600-700 Million people,** which are roughly 8%-10% of the 7 Billion+ world population.

### Software Solutions for Field Service Scheduling
This section describes several software solutions to support a service scheduler in his job; each of which represents a group of products available on today's market. You'll intuitively detect different levels of intelligence, which we will later put into a structured paradigm. We start by describing only the functionalities of four software products, and then proceed to discuss their intelligence by means of decision-making capabilities and automation.
**Software D:** Computer serves solely as an electronic scheduling board (Gantt chart) replacing metal board with physical magnetic tiles. The user decides on job assignments (who, what, where, when, …)  and places them on the board. No software services are offered beyond graphic display and KPI calculation. Still it is more efficient than doing it on paper or a metal board, thus business value is generated.
**Software C**: The user decides on job schedules, places them on the electronic graphic board, and the software comments on business rules violations, e.g. violations of SLA commitments, or insufficient skills of the field engineer. The software acts as a "Responsive board" commentator, but that's all, and without any automatic decision-making capabilities. It serves as a 'watch dog' passive advisor. It does contribute, however, to speeding up the solution process and to its quality, i.e. there is business value in terms of business results. Software C is capable of judging whether a solution is valid or not, and calculating its value, but it is not able to generate solutions.
**Software B:** In addition to what Software C offers, by user touching an individual service job, Software B proposes candidate assignment options (who, when…) that comply with the business rules and are ranked by a fitness score. The user then selects the preferred assignment, and it gets displayed on the electronic board. The **sequence** of service jobs is decided by the user. If dead-ends are reached, it is the user who decides which assignments to lift off the board (backtracking) and resume the process.
**Software A**: The user selects a group of jobs (daily/weekly/monthly) and clicks **SCHEDULE.** Within seconds or minutes, the software, **fully automatically,**

**generates the complete schedule in a way that complies with the business rules and its quality score is considerably higher than what a human would have produced**.

### Discussion of Software Solutions for Field Service Scheduling

Software D represents traditional data processing software used by a human scheduler where the computer is used merely as a graphic, recording, and calculating aid, and has no clue of the domain concepts or what's behind the numbers it calculates. Typically, with Software D it takes longer (relative to C, B, A) to build a full schedule and to manage it throughout the day. Most important, **schedule quality is a function of the skills, experience, and knowledge of the individual human scheduler in the territory or business line**. This dependence on human's skills, time and brain power is critical, as the schedules produced, day in and day out, eventually impact the quarterly financial metrics of the company, and hence its share price.

Most of us are likely to agree that Software C has a higher level of smartness than Software D, and that the overall schedule quality with C is likely to be higher than in D. In fact, our experience shows that with software D, even when the user is an experienced scheduler, human-generated schedule contains non-negligible number of rule violations. Software C however, does not contribute much to speed up the process, i.e. automation.

With Software B, we see an increase in the level of smartness and level of automation, but both are still relatively low, consuming significant time of the scheduler. It is **the user who drives the solution process**, makes the decisions on each individual assignment and backtracks from dead-ends. This makes scheduling a labor-intensive job, and, in a typical service organization, one scheduler can only manage 10 to 15 field engineers. Assume 100 schedulers for 1,500 field engineers at an annual labor cost of $75,000 per scheduler, and the total labor cost for scheduling amounts to $7.5M, annually. If we can increase substantially the 1:10 to 1:15 ratio, we can save considerably on labor cost. This takes us to Software A.

With software A, we reach very high levels on both dimensions: full automation at high speed, and superb schedule quality with no rule violations. With Software A, **the problem-solving process is not dependent on instructions from the user; it is fully autonomous**. Yet, **the user has the option to modify, override any of the algorithm's decisions, and the algorithm will continue to offer its services from the post-intervention point**. The business value is quantifiable in terms of both labor cost of human schedulers, and operational efficiencies. As an example, in a Canadian utility company, once ClickSchedule was put to regular use, the number of dispatchers was reduced from 75 to 25, saving 65% on labor cost (est. $ 2.5M a year). The AI optimization algorithm also managed to squeeze an average of 4,800 jobs daily, up from 4,000 with the manual software, yielding 20% productivity growth and an increase of about $5m a year in service value.

### AI Quadrant (AIQ)

The proposed AI-Quadrant (fig.1) achieves nicely the goal of obviating the need to develop binary criteria for a "non-AI software" category: score any software on the two dimensions, position it on the AI Quadrant and let the viewer decide. A low score on one or both dimensions speaks for itself. The narrow stripes to the right of the vertical axis and just above the horizontal axis represent the "**Borderline AI**" region. Indeed, these are the regions where you still the need heavy involvement of a skilled experienced human, either because of lack of automation, or lack of smartness, and therefore a software in these regions has minimal business value. Figure 1 also shows AA and AAA software, which will be introduced and discussed in subsequent sections.

Before we proceed, let us point at one more type of software, type F.

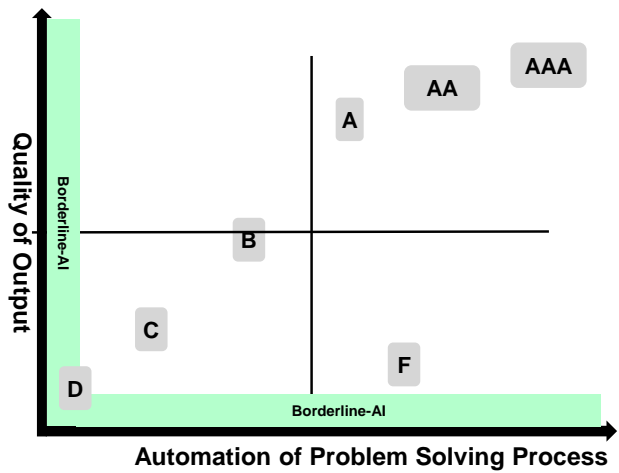

Figure 1: AI Quadrant

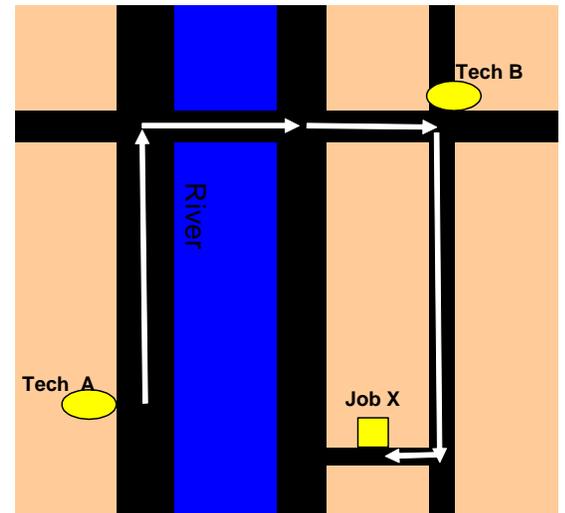

Figure 2: Glaring Mistake Due to Air-Distance Assumption

### On Glaring Mistakes and High Automation Level

**Software F (Failure):** Consider a scheduling software for field service where the distance factor for job assignment considerations is calculated by air-distance; as opposed to street level driving. Such an algorithm may assign (Fig 2) Job X to Tech A whose air-distance location is ½ mile across the river, and not to B who is three miles away on the same side of the river as customer X. This is a very poor decision, considering that Tech A needs to drive 3 miles up the river to the nearest bridge, and then 3 miles down the river to Customer X's location. Despite the high level of automation of Software F, no user will accept software with "discontinuities" in its intelligence whereby it occasionally makes glaring mistakes that no human would make, i.e. very low score on the Quality dimension. This typically happens when the software developer forces a real-life problem into the straight jacket of a mathematical model (air-distance) in order to make the algorithm works in practical time. This is like forcing linear models on a real-life problem which inherently involves complex non-linear relationships. Software products type F were one of the factors contributing to the "AI Winter" of the 1990's and early 2000's.

### Machine Learning Elevates Intelligence Level to AA

No AI discussion is complete without discussing recent years' remarkable breakthroughs in Machine Learning (ML) algorithms and their outstanding results. With IoT and Big Data technologies, ML algorithms can certainly improve the quality of solutions produced by AI software, as well as their automation level, thereby increasing their business value. For example, ML algorithms can use historic data to learn statistical patterns of "task duration" that integrate a variety of factors including the individual technician, job type, territory, and time of year. More accurate estimates of task duration produce higher quality schedules, elevating an A level software to **AA** level (Fig. 1).

### Conversational Intelligent Digital Assistants Elevate AI Score to AAA level

Business AI software that covers a rich spectrum of business workflows with high levels of smartness and automation enables AI-based digital assistants that shadow the user throughout the day and help him manage the **dynamics of the day.** The key characteristics of such AI software include:

1. **Anticipates** user needs and acts on them **proactively,** within the boundaries delegated to it. It does not wait for the user to ask or instruct.
2. Constantly **monitors** the states of all objects in the applications space, and accordingly generates alerts.
3. **Recognizes context change automatically**,
4. Infers whether there is a **need for action**, and

5. Automatically generates **action-oriented decisions**, with possible automatic execution within the degrees of authority it was configured for, e.g. close the valve to stop radioactive leak.

Intelligent personal assistants with such abilities have been part of the W6 solutions since the 1990's. We named them **Butlers** as an analogy to the classic English butler; the wise personal assistant who is always one step ahead of his master and takes care of his needs even before the master is aware of them. See for example a 2013 integration with IBM mobility for workforce applications [6], and [3].

In a subsequent paper [3] I cover with greater detail **conversational intelligent digital assistants for business roles**. These hand-free digital assistants leverage the recent progress in speech and natural language understanding with domain-specific decision-making algorithms. They are **likely to be the next generation of today's conversational Bots**. For example, Wired Magazine (September 2018) reports a story [5] where existing intelligent digital assistant for factory floor decision-making by Plataine is integrated with the current version of Google Glass and its Dialogflow voice-interface service. A sample message that it proactively and automatically generated sounds like this: *"We have just been informed that composite material Roll A784 is defected. We manufactured the following 15 kits from this roll (displayed on Glass) which we now need to remake. The specific rolls that need to be pulled from the storage room are also included taking into consideration expiration dates and remnants length. Upon your approval I will also automatically produce the nesting, cutting, and assembly plan".* The user may just say:" OK" to approve as is or may modify the plan after some what-if simulations. In any case, managing the situation with close to optimal decisions is done in minutes not hours. The architecture of Plataine's digital assistants is designed to behave by the five key characteristics listed above. Sounds simple to implement? Not quite. In fact, to achieve such behavior you need multiple algorithms from different AI areas, e.g. **pattern recognition** algorithms to automatically recognize contexts, **prediction** algorithms to assess where current context might evolve to a point when an action is needed, and **search/optimization** algorithms to decide on the best action(s) for the current context. With such AI software, we further improve the spectrum and quality of business decisions and increase the level of automation, leading to overall higher business value, taking the AI product's intelligence score to **AAA**.

### Summary Comments

Since the focus of the article is on business applications and our fundamental assumption is that a higher level of software intelligence implies higher value to its users, it makes sense to have separate AI Quadrants for different business functions. Similar to the quadrants that industry research firms like Gartner and Forrester have for different business functions, e.g. ERP, CRM, HR. The proposed AI-Quadrant's dimensions (smartness and automation), however, are universal, and I believe can also be used for building AI-Quadrants for more generic AI applications such as Image, Speech, or Natural Language Understanding.

**[* Moshe BenBassat:](#)** *For several decades Moshe BenBassat has been researching, practicing and educating* **Artificial Intelligence**; *from the first AI "Spring" of the 1980's, during the AI "Winter" of the 1990th and early 2000 years, and now in the new AI "Spring" of the 21th century. During a long academic career with positions at Tel Aviv University, USC, and UCLA, Professor BenBassat made significant contributions in Pattern Recognition, Artificial Intelligence, Optimization, Data Science and Machine Learning. Following his invention of* **"service chain optimization"** *(patent awarded), he founded* **ClickSoftware** *which have been leading this space since its inception in 1997. Moshe served as ClickSoftware's CEO until 2015, at which point it was acquired by a private equity firm. Moshe also founded* **Plataine** *which is focused* **on intelligent automation for smart manufacturing, leveraging Industrial Internet of Things (IIoT) and Artificial Intelligence** *technologies.*